\documentclass[10pt, a4paper]{article}
\usepackage{lrec2022} 
\usepackage{multibib}
\newcites{languageresource}{Language Resources}
\usepackage{graphicx}
\usepackage{tabularx}
\usepackage[normalem]{ulem}
\usepackage{colortbl}
\usepackage{soul}
\usepackage[table,xcdraw]{xcolor}

\usepackage{titlesec}
\titleformat{\section}{\normalfont\large\bfseries\center}{\thesection.}{1em}{}
\titleformat{\subsection}{\normalfont\SmallTitleFont\bfseries\raggedright}{\thesubsection.}{1em}{}
\titleformat{\subsubsection}{\normalfont\normalsize\bfseries\raggedright}{\thesubsubsection.}{1em}{}
\renewcommand\thesection{\arabic{section}}
\renewcommand\thesubsection{\thesection.\arabic{subsection}}
\renewcommand\thesubsubsection{\thesubsection.\arabic{subsubsection}}

\usepackage{epstopdf}
\usepackage[utf8]{inputenc}

\usepackage{hyperref}
\usepackage{xstring}
\usepackage{makecell}
\usepackage{color}
\usepackage{arydshln}
\usepackage{xcolor}
\usepackage{float}
\usepackage{stfloats}

\title{Annotating the Tweebank Corpus on Named Entity Recognition and Building NLP Models for Social Media Analysis}

\name{Hang Jiang\thanks{The first two authors contribute equally. YH is also affiliated with Harvard Medical School.}, Yining Hua, Doug Beeferman, Deb Roy} 

\address{MIT Center for Constructive Communication\\
 75 Amherst St, Cambridge, MA 02139 \\
 \{hjian42, ninghua, dougb5, dkroy\}@mit.edu\\
 }

\abstract{Social media data such as Twitter messages (``tweets'') pose a particular challenge to NLP systems because of their short, noisy, and colloquial nature. Tasks such as Named Entity Recognition (NER) and syntactic parsing require highly domain-matched training data for good performance. To date, there is no complete training corpus for both NER and syntactic analysis (e.g., part of speech tagging, dependency parsing) of tweets. While there are some publicly available annotated NLP datasets of tweets, they are only designed for individual tasks. In this study, we aim to create Tweebank-NER, an English NER corpus based on Tweebank V2 (TB2), train state-of-the-art (SOTA) Tweet NLP models on TB2, and release an NLP pipeline called Twitter-Stanza. We annotate named entities in TB2 using Amazon Mechanical Turk and measure the quality of our annotations. We train the Stanza pipeline on TB2 and compare with alternative NLP frameworks (e.g., FLAIR, spaCy) and transformer-based models. The Stanza tokenizer and lemmatizer achieve SOTA performance on TB2, while the Stanza NER tagger, part-of-speech (POS) tagger, and dependency parser achieve competitive performance against non-transformer models. The transformer-based models establish a strong baseline in Tweebank-NER and achieve the new SOTA performance in POS tagging and dependency parsing on TB2. We release the dataset and make both the Stanza pipeline and BERTweet-based models available ``off-the-shelf'' for use in future Tweet NLP research. Our source code, data, and pre-trained models are available at: \url{https://github.com/social-machines/TweebankNLP}.
 \\ \newline \Keywords{text annotation, noisy text, NLP toolkit, Twitter, named entity recognition, tokenization, lemmatization, part-of-speech tagging, dependency parsing}}
 
\begin{document}

\maketitleabstract

\section{Introduction}

Researchers use text data from social media platforms such as Twitter and Reddit for a wide range of studies including opinion mining, socio-cultural analysis, and language variation. Messages posted to such platforms are typically written in a less formal style than what are found in conventional data sources for NLP models, namely news articles, papers, websites, and books. Processing the noisy and informal language of social media is challenging for traditional NLP tools because such messages are usually short in length and irregular in spelling and structure. In response, the NLP community has been constructing language resources and building NLP pipelines for social media data, especially for Twitter.


Annotating social media language resources is important to the development of NLP tools. \newcite{foster2011hardtoparse} is the one of the earliest attempts to annotate tweets in the Penn Treebank (PTB) format. Following a similar PTB-style convention suggested by \newcite{schneider2013framework}, \newcite{kong2014dependency} created Tweebank V1. However, the PTB annotation guidelines leave many annotation decisions unspecified and are therefore unsuitable for informal and user-generated text. After Universal Dependencies (UD) \cite{nivre2016universal} was introduced to enable consistent annotation across different languages and genres, \newcite{liu2018parsing} introduced a new tweet-based Tweebank V2 in UD, including tokenization, part-of-speech (POS) tags, and (labeled) Universal Dependencies. Besides syntactic annotation, NLP researchers have also annotated tweets on named entities. \newcite{ritter2011named} first introduced this English Twitter NER task and found that NER systems trained on the news perform poorly on tweets.  Since then, the noisy user-generated text (WNUT) workshop has proposed a few benchmark datasets including WNUT15 \cite{ws-2015-noisy}, WNUT16 \cite{ws-2015-noisy}, and WNUT17 \cite{ws-2017-noisy} for Twitter lexical normalization and named entity recognition (NER). However, these benchmarks are not based upon TB2, which contains high-quality UD annotations. 
Annotating named entities in TB2 fills a gap in NLP research, allowing researchers to train multi-task learning models in NER, POS tagging, and dependency parsing, and study the linguistic relationship between syntactic labels and named entities in the Twitter domain.

Many researchers have invested in building better NLP pipelines for tokenization, POS tagging, parsing, and NER. The earliest work focuses on Twitter POS taggers \cite{gimpel2010part,owoputi2013improved} and NER \cite{ritter2011named}. Later, \newcite{kong2014dependency} published TweeboParser on Tweebank V1 to include tokenization, POS tagging, and dependency parsing. \newcite{liu2018parsing} further improved the whole pipeline based on TB2. The current state-of-the-art (SOTA) pipeline in POS tagging and NER is based on BERT pre-trained on a large number of tweets \newcite{nguyen2020bertweet}. However, these efforts (1) are often no longer maintained \cite{ritter2011named,kong2014dependency}, (2) do not contain publicly available NLP models (e.g., NER, POS tagger) \cite{nguyen2020bertweet}, (3) are written in C/C++ or R with complicated dependencies and installation process (e.g., Twpipe \cite{liu2018parsing} and UDPipe \cite{straka2016udpipe}), making them difficult to be integrated into Python frameworks and to be used in an ``off-the-shelf'' fashion.  Many modern NLP tools in Python such as spaCy\footnote{\url{https://spacy.io/}}, Stanza \cite{qi2020stanza}, and FLAIR \cite{akbik2019flair} have been developed for standard NLP benchmarks but have never been adapted to Tweet NLP tasks. In this study, we choose Stanza over other NLP frameworks because (1) the Stanza framework achieves SOTA or competitive performance on many NLP tasks across 66 languages \cite{qi2020stanza}, (2) Stanza supports both CPU and GPU training and inference while transformer-based models (e.g., BERTweet) need GPU, (3) Stanza shows superior performance against spaCy in our experiments despite slower speeds, (4) Stanza is competitive in speed compared with FLAIR of similar accuracy \cite{qi2020stanza}, but the FLAIR dependency parser is still under development.


In this paper, we annotate Tweebank V2 on NER to create \texttt{Tweebank-NER} and also build Tweet NLP models based on Stanza and transformer models.  We run additional experiments to answer the following questions: (1) How is the quality of the NER annotations? (2) Do NER models trained on existing Twitter NER data perform well on \texttt{Tweebank-NER}? (3) How do Stanza models perform compared with other NLP frameworks on the core Tweet NLP tasks? (4) How do transformer-based models perform compared with traditional models on these tasks? Our contributions are as follows:

\begin{itemize}
\setlength\itemsep{0em}
\item We annotate Tweebank V2, the main treebank for English Twitter NLP tasks, on NER. This annotation not only provides a new benchmark (\texttt{Tweebank-NER}) for Twitter NER but also makes Tweebank a complete dataset for both syntactic tasks and NER, making it suitable for training multi-task learning models in POS tagging, dependency parsing, and NER.
\item We leverage the Stanza framework to present an accurate and fast Tweet NLP pipeline called \verb|Twitter-Stanza|. It includes NER, tokenization, lemmatization, POS tagging, and dependency parsing modules, and it supports both CPU and GPU computation. 

\item We compare \verb|Twitter-Stanza| against existing models for each presented NLP task, confirming that Stanza's simple neural architecture is effective and suitable for tweets. Among non-transformer models, the \verb|Twitter-Stanza| tokenizer and lemmatizer achieve SOTA performance on TB2, and its POS tagger, dependency parser, and NER model obtain competitive performance. 


\item We also train transformer-based models to establish a strong baseline on the \texttt{Tweebank-NER} benchmark and SOTA performance in POS tagging and dependency parsing on TB2. We upload the BERTweet-based NER and POS taggers to the Hugging Face Hub: \url{https://huggingface.co/TweebankNLP}

\item We release our data, models, and code. Our \verb|Twitter-Stanza| pipeline is highly compatible with Stanza's Python interface and is simple to use in an ``off-the-shelf'' fashion. We hope that our \verb|Twitter-Stanza| and Hugging Face BERTweet models can serve as a convenient NLP tool and a strong baseline for future research and applications of Tweet analytic tasks.

\end{itemize}





\section{Dataset and Annotation Scheme}
In this study, we primarily work on the Tweebank V2 dataset and develop its NER annotations through rigorous annotation guidelines. We also evaluate the quality of our annotations, showing that it has a good F1 inter-annotator agreement score. 


\subsection{Datasets and Annotation Statistics}
Tweebank V2 (TB2) \cite{kong2014dependency,liu2018parsing}is a collection of 3,550 labeled anonymous English tweets annotated in Universal Dependencies. It is a commonly used corpus for the training and fine-tuning of NLP systems on social media texts.  A summary of TB2 is shown in Table \ref{tab:stats}.

\begin{table}[!ht]
\centering
\begin{tabular}{l|ccc}
\textbf{Dataset}& \textbf{Train} & \textbf{Dev} & \textbf{Test}\\ \Xhline{2\arrayrulewidth}
Tweets & 1,639 & 710 & 1,201 \\
Tokens& 24,753 & 11,742 & 19,112\\
Avg. token per tweet& 15.1 & 16.6 & 15.9 \\
Annotated spans & 979 & 425 & 750 \\ 
Annotated tokens& 1,484 & 675 & 1183\\ 
Avg. token per span & 1.5 & 1.6 & 1.6\\ 
\end{tabular}
\caption{Annotated corpus statistics.}
\label{tab:stats}
\end{table}

\subsection{Annotation Guidelines}
We follow the CoNLL 2003 guidelines\footnote{\url{https://www.clips.uantwerpen.be/conll2003/ner/}} to annotate named entities. We are aware that some NER annotations (e.g., English OntoNotes) have more than four classes. We adopt the standard four-class CoNLL 2003 NER guidelines for two reasons. One one hand, adopting a more fine-grained annotation scheme is more challenging for human annotators. The 4-class scheme is already quite challenging for humans since the inter-annotator agreement is low for the MISC class. On the other hand, Tweebank is relatively small, with only 3,550 tweets. An annotation scheme with more classes than that will mean fewer instances per class, and greater difficulty for NER models to learn efficiently. To help annotators understand the guidelines, we provide multiple examples for each rule and ask annotators to read them before the task. Our task focuses on the following four named entities:
\begin{itemize}
\setlength\itemsep{0em}
\item \textbf{PER}: persons (e.g., Joe Biden, joe biden, Ben, 50 Cent, Jesus)
\item \textbf{ORG}: organizations (e.g., Stanford University, stanford, IBM, Black Lives Matter, WHO, Boston Red Sox, Science Magazine, NYT) 
\item \textbf{LOC}: locations (e.g., United States, usa, China, Boston, Bay Area, CA, MT Washington)
\item \textbf{MISC}: named entities which do not belong to the previous three. (e.g., Chinese, chinese, World Cup 2002, Democrat, Just Do It, Top 10, Titanic, The Shining, All You Need Is Love)
\end{itemize}

To handle challenges in tweets, we also add requirements consistent with \cite{ritter2011named}: (1) ignore numerical entities (MONEY, NUMBER, ORDINAL, PERCENT), (2) ignore temporal entities (DATE, TIME, DURATION, SET), (3) "At mentions" are not named entities (e.g., allow ``Donald Trump'' but not @DonaldTrump), (4) \#hashtags are not named entities (e.g., allow ``BLM'' but not ``\#BLM''), (5) URLs are not named entities (e.g., disallow https://www.google.com/). 

\subsection{Annotation Logistics}
We use the Qualtrics platform to design the sequence labeling task and Amazon Mechanical Turk to recruit annotators. We first launch a pilot study, annotate each of the 100 tweets, and discuss tweets with divergent annotations. Based on the pilot study, we develop a series of annotation rules and precautions. During the recruiting process, each annotator is given an overview of annotation conventions and our guidelines, after which they are asked to complete the qualification test. The qualification test consists of 7 tweets that are selected from the pilot study. 
An annotator must make fewer than 2 errors and not make any significant error in order to pass the qualification test.  We consider a significant error to be one
in which any URL, @USER, or hashtag is labeled as a named entity;  or one in which the PERSON, LOCATION, and ORG categories are confused with each other.

After all tweets have been annotated by at least 3 annotators, we merge the annotation results and create the \texttt{Tweebank-NER} dataset in the BIO format \cite{ratinov2009design}. In the merging process, if at least two annotators give the annotation result for a tweet, we use that result as the final annotation. Otherwise, we discuss and re-annotate the tweet to reach a consensus. We identify 178 span annotations whose three annotations are different from each other and decide their gold annotations collectively by two authors.
We find that one of the three annotators' answers is the same as the final annotation for 155 out of the 178 annotations.



\subsection{Annotation Quality}
We first evaluate the quality of the annotations using a measure of inter-annotator agreement (IAA). For NER, Cohen's Kappa is not the best measure because it needs the number of negative cases, but NER is a sequence tagging task. Therefore, we follow previous work \cite{hripcsak2005agreement,grouin2011proposal,brandsen2020creating} to use the token-level pairwise F1 score calculated without the O label as a better measure for IAA in NER \cite{deleger2012building}. In Table \ref{tab:iaa}, we observe that PER, LOC, and ORG have higher F1 agreement than MISC, showing that MISC is more difficult to annotate than the other classes. We also provide the additional Kappa measure ($\kappa=$0.347) on annotated tokens to provide some insights, although it significantly underestimates IAA for NER. Finally, we calculate the scores by comparing the crowdsourced annotators against our own internal annotations on 100 sampled examples, obtaining a similar F1 score (0.71). 

\begin{table}[!ht]
\centering
\begin{tabular}{l|cc}
\textbf{Label}& Quantity & F1\\ \Xhline{2\arrayrulewidth}
PER & 777 & 84.6\\
LOC & 317 & 74.4 \\
ORG & 541 & 71.9 \\
MISC& 519 & 50.9 \\ 
\hline
Overall & 2,154 & 70.7
\end{tabular}
\caption{Number of span annotations per entity type and Inter-annotator agreement scores in pairwise F1.}
\label{tab:iaa}
\end{table}

We analyzed the 178 annotations passed to the merge step, finding that the proportion of each label is 8.4\% (LOC), 15.2\% (PER), 29.2\% (ORG), and 47.2\% (MISC). These numbers show that MISC is the most challenging class for human annotators and ORG is also relatively difficult compared to LOC and PER. This confirms the IAA measured in pairwise F1 in Table \ref{tab:iaa} because the MISC has the lowest F1 (50.9\%) and ORG has the second lowest F1 (71.9\%). 


In the future, we suggest a few ways to improve the annotation quality.  The first way is to increase the number annotators per tweet in both the initial and merge stages. Second, hiring a small number of experienced annotators instead of using crowdsourcing platforms will make the annotations more consistent. Third, adopting a human-in-the-loop approach allows annotators to focus on difficult instances from MISC and ORG, which can reduce the cost and improve the performance of the models at the same time. 



\section{Methods for NLP Modeling}

Stanza is a state-of-the-art and efficient framework for many NLP tasks \cite{qi2020stanza,zhang2021biomedical} and it supports both NER and syntactic tasks. We use Stanza to train NER models as well as syntactic models (tokenization, lemmatization, POS tagging, dependency parsing) on TB2. For more detailed information on Stanza, we refer the readers to the Stanza paper \cite{qi2020stanza} and its current website\footnote{\url{https://stanfordnlp.github.io/stanza/}}. We use Twitter GloVe embeddings \cite{pennington2014glove} with 100 dimensions in our experiments and the default parameters in Stanza for training. 

Alternative NLP frameworks such as spaCy, FLAIR, transformers, and spaCy-transformers are compared with Stanza. Both spaCy and FLAIR are open-source NLP frameworks for NER and syntactic tasks. Transformers is a library of pre-trained transformer models for NLP and it provides a TokenClassification module\footnote{\url{https://github.com/huggingface/transformers/tree/main/\\examples/legacy/token-classification}}, which is adopted for NER and POS tagging. We denote these models as HuggingFace-BERTweet in our experiments. The spaCy-transformers framework provides the spaCy interface to combine pre-trained representations from transformer-based language models and its own NLP models via Hugging Face's transformers. To train spaCy, we adopt the default NER setting\footnote{\url{https://github.com/explosion/projects/blob/v3/pipelines/\\ner\textunderscore wikiner/configs/default.cfg}} and the default syntactic NLP pipeline\footnote{\url{https://github.com/explosion/projects/tree/v3/\\benchmarks/ud\textunderscore benchmark}}. For FLAIR, we train its NER and syntactic modules with the default settings as well. For spaCy-transformers models, we finetune BERTweet-base and XLM-RoBERTa-base language models via spaCy-transformers for NER, POS Tagging, and dependency parsing\footnote{\url{https://github.com/explosion/projects/blob/v3/benchmarks/\\ud\_benchmark/configs/transformer.cfg}}. We denote them as spaCy-BERTweet and spaCy-XLM-RoBERTa in the paper. BERTweet \cite{nguyen2020bertweet} is the first public large-scale language model for English tweets based on RoBERTa and XLM-RoBERTa-base is a multilingual version of RoBERTa-base. All transformer-based models show strong performance in Tweet NER and POS tagging \cite{nguyen2020bertweet}. The architecture and training details of the models above can be found at our public repository.


\subsection{Named Entity Recognition}
In this paper, we adopt the four-class convention to define NER as a task to locate and classify named entities mentioned in unstructured text into four pre-defined categories: PER, ORG, LOC, and MISC \cite{sang2003introduction}. We use the Stanza NER architecture for training and evaluation, which is a contextualized string representation-based sequence tagger \cite{akbik2018contextual}. This model contains a forward and a backward character-level LSTM language model to extract token-level representations and a BiLSTM-CRF sequence labeler to predict the named entities. We also train the default NER models for SpaCy, FLAIR, HuggingFace-BERTweet, and spaCy-BERTweet for comparison.




\subsection{Syntactic NLP Tasks}
\subsubsection{Tokenization}

Tokenizers predict whether a given character in a sentence is the end of a token. The Stanza tokenizer jointly works on tokenization and sentence segmentation, by modeling them as a tagging problem over character sequences.  In accordance with previous work \cite{gimpel2010part,liu2018parsing}, we focus on the performance in tokenization, as tweets are usually short with a single sentence. 

To compare with spaCy, we train a spaCy tokenizer named char\_pretokenizer.v1. FLAIR uses spaCy's tokenizer, so we exclude it from comparison. We also include baselines mentioned in previous work \cite{kong2014dependency,liu2018parsing}. Twokenizer \cite{o2010tweetmotif} is a regex-based tokenizer and does not adapt to the UD tokenization scheme. Stanford CoreNLP \cite{manning2014stanford}, spaCy, and UDPipe v1.2 \cite{straka2017tokenizing} are three popular NLP frameworks re-trained on TB2. Twpipe tokenizer \cite{liu2018parsing} is similar to UDPipe, but replaces GRU in UDPipe with an LSTM and uses a larger hidden unit number. We do not compare with transformer-based models because they use subword-level tokenization schemes like WordPiece \cite{wu2016google} and BPE \cite{sennrich2015neural}. 


\subsubsection{Lemmatization}

Lemmatization is the process of recovering each word in a sentence to its canonical form. We train the Stanza lemmatizer on TB2, which is implemented as an ensemble model of a dictionary-based lemmatizer and a neural seq2seq lemmatizer. We compare the Stanza lemmatizer against three lemmatizers from spaCy, NLTK, and FLAIR (Table \ref{tab:lemmatization}). Both NLTK and spaCy lemmatizer are rule-based and use a dictionary to look up the canonical form given a word and it POS tag. The FLAIR lemmatizer is a char-level seq2seq model. We provide gold POS tags for lemmatization.

\subsubsection{POS Tagging} 
POS tagging assigns each token in a sentence a POS tag. We train the Stanza POS tagger, a bidirectional long short-term memory network as the basic architecture to predict the universal POS (UPOS) tags. We ignore the language-specific POS (XPOS) tags because TB2 only contains UPOS tags. 

We also train the default POS taggers for SpaCy, FLAIR, HuggingFace-BERTweet, spaCy-BERTweet, spaCy-XLM-RoBERTa. We include performance from existing work in Tweet POS tagging: (1) Stanford CoreNLP tagger, (2) \newcite{owoputi2013improved}'s word cluster–enhanced greedy tagger, (3) \newcite{owoputi2013improved}'s word cluster–enhanced tagger with CRF, (4) \newcite{ma2016end}'s neural tagger, (5) BERTweet-based POS tagger \cite{nguyen2020bertweet}. The first four models were re-trained on the combination of TB2 and UD\_English-EWT \citelanguageresource{ewt} training sets, whereas the BERTweet-based tagger was fine-tuned solely on TB2. HuggingFace-BERTweet has the same architecture implementation as \newcite{nguyen2020bertweet}.

\subsubsection{Dependency Parsing}

Dependency parsing predicts a syntactic structure for a sentence, where every word in the sentence is assigned a syntactic head that points to either another word in the sentence or an artificial root symbol. Stanza's dependency parser combines a Bi-LSTM-based deep biaffine neural parser \cite{dozat2016deep} and two linguistic features, which can significantly improve parsing accuracy \cite{qi2019universal}. Gold-standard tokenization and automatic POS tags are used.

We also re-train spaCy, spaCy-BERTweet, and spaCy-RoBERTa dependency parsers with their default parser architectures\footnote{FLAIR and Hugging Face's transformers do not contain dependency parsing by default. }. We compare our Stanza models with previous work: (1) \newcite{kong2014dependency}'s graph-based parser with lexical features and word cluster and it uses dual decomposition for decoding, (2) \newcite{dozat2016deep}'s neural graph parser with biaffine attention, (3) \newcite{ballesteros2015improved}'s neural greedy stack LSTM parser, (4) an ensemble model of 20 transition-based parsers \cite{liu2018parsing}, (5) A distilled graph-based parser of the previous ensemble model \cite{liu2018parsing}. These models are all trained on TB2+UD\_English-EWT. We are aware that \newcite{stymne2020cross} trained a transition-based uuparser \cite{de2017raw} on a combination of TB2, UD\_English-EWT, and more out-of-domain data (English GUM \cite{Zeldes2017}, LinES \cite{ahrenberg-2007-lines}, ParTUT \cite{sanguinetti2015parttut}) to further boost model performance, but we do not experiment with this data combination to be consistent with \newcite{liu2018parsing}.

\section{Evaluation}
We train the NER and syntactic NLP models described above with 1) TB2 training data (the default data setting), 2) TB2 training data + extra Twitter data (the combined data setting). For the combined data setting, we add the training and dev sets from other data sources to TB2's training and dev sets respectively. Specifically, we add WNUT17\footnote{We map both ``group'' and ``corporation'' to ``ORG'', and both ``creative work'' and ``product'' to ``MISC''. }  \cite{ws-2017-noisy} for NER. For syntactic NLP tasks, we add UD\_English-EWT \citelanguageresource{ewt}. We pick the best models based on the corresponding dev sets and report their performance on their TB2 test sets. For each task, we compare Stanza models with existing studies and alternative NLP frameworks. 

\subsection{Performance in NER}

\begin{table}[h]
\centering
\begin{tabular}{l|cc}
\textbf{Systems} & \textbf{F1}\\ \Xhline{2\arrayrulewidth}
spaCy (TB2) & 52.20\\
spaCy (TB2+W17) & 53.89\\
FLAIR (TB2) & 62.12\\
FLAIR (TB2+W17) & 59.08 \\
\hdashline
HuggingFace-BERTweet (TB2) & 73.71 \\
HuggingFace-BERTweet (TB2+W17) & \textbf{74.35} \\
spaCy-BERTweet (TB2) & 73.79 \\
spaCy-BERTweet (TB2+W17) & 74.15 \\
\hline
Stanza (TB2) & 60.14 \\
Stanza (TB2+W17) &62.53
\end{tabular}
\caption{NER comparison on the TB2 test set in entity-level F1. ``TB2'' indicates to use the TB2 train set for training. ``TB2+W17'' indicates to combine TB2 and WNUT17 train sets for training. }
\label{tab:NER}
\end{table}

\subsubsection{Main Findings}
The NER experiments presented in Table \ref{tab:NER} show that the Stanza NER model (TB2+W17) achieves the best performance among all non-transformer models. At the same time, the Stanza model is up to 75\% smaller than the second-best FLAIR model \cite{qi2020stanza}. For transformer-based approaches, spacy-BERTweet and HuggingFace-BERTweet have close performance to each other. The HuggingFace-BERTweet approach trained on TB2+W17 achieves the highest performance (74.35\%) on \texttt{Tweebank-NER}, establishing a strong benchmark for future research. We also find that combining the training data from both WNUT17 and TB2 improves the performance of spaCy, FLAIR, Stanza, and BERTweet-based models.



\begin{figure}[ht]
\begin{center}
\includegraphics[scale=0.45]{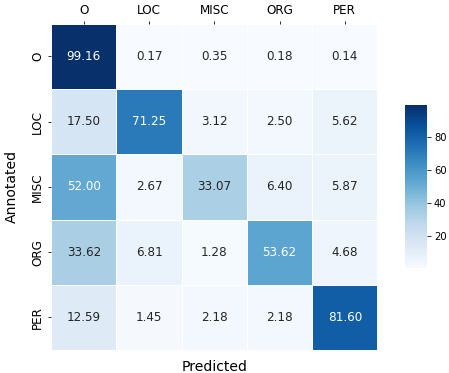} 
\end{center}
\caption{Confusion matrix generated by the Stanza (TB2+W17) model to show percentages for each combination of predicted and true entity types.}
\label{fig:ner}
\end{figure}

\subsubsection{Confusion Matrix Analysis}
In Figure \ref{fig:ner}, we plot a confusion matrix for all four entity types and ``O'', the label for tokens that do not belong to any of these types.  The diagonal and the vertical blue lines are expected because the cells on the diagonal are when the algorithm predicts the correct entity and the vertical line is when the algorithm mistakes an entity for the ``O'' entity, which is the most common error for NER. We notice that MISC entities are easily mistaken as ``O'', which corresponds to the annotation statistics in Table \ref{tab:iaa}, where MISC has the lowest IIA score in pairwise F1. Thus, MISC is the most challenging of the four types for both humans and machines.

\definecolor{ao(english)}{rgb}{0.0, 0.5, 0.0}
\begin{table*}[ht]
\centering
\begin{tabular}{l|lll}
\textbf{Error type}                   & \textbf{ weet example} \\ \Xhline{2\arrayrulewidth} 
PER $\rightarrow$ O                & The 50 \% Return Method Billionaire Investor \textbf{\textcolor{red}{Warren Buffet}} Wishes He Could Use &  &  \\

LOC $\rightarrow$ O                & Getting ready ... @ \textbf{\textcolor{ao(english)}{Pasco Ephesus Seventh} \textcolor{red}{- day Adventist Church}}                       &  &   \\

ORG $\rightarrow$ O                & \#bargains \#deals 10.27.10 \textcolor{red}{\textbf{Guess Who}} `` American Woman '' Guhhh deeeh you !                             &  &   \\

MISC $\rightarrow$ O & RT @USER1508 : Do you ever realize \textcolor{red}{\textbf{Sounds Live Feels Live}} Starts this month and just                                       &  &   
\end{tabular}
\caption{Common mistakes made by the Stanza (W17+TB2) NER model for each error type. ``X $\rightarrow$ O'' means the model predicts X entity to be O by mistake. Green and red texts are gold annotations of the corresponding type in each row. Correct predictions are in bold green and gold annotations missed by the model are in bold red.}
\label{tab:examples}
\end{table*}

\subsubsection{Error Analysis}
We identify the most common error types that Stanza (TB2+W17)\footnote{We pick Stanza over BERTweet for error analysis because we only aimed to publish the Stanza pipeline at the beginning. We eventually publish the BERTweet models too.} makes on the TB2 test in Figure \ref{fig:ner}: predicting PER, LOC, ORG, MISC to be O.  We pick some representative examples for each error type, shown in Table \ref{tab:examples}.  For the  $PER\rightarrow O$ error type, every first letter in a word is capitalized and the model fails to recognize the famous investor ``Warren Buffet'' in such a context. We find that person entities with abbreviations (e.g., ``GD'' for ``G-dragon''), lower case (e.g., ``kush'' for ``Kush''), or irregular contextual capitalization are challenging to the NER system. For the  $LOC\rightarrow O$ error type, the  structure to encode location is complicated and sometimes interrupted by the parentheses and dashes (e.g., ``- day Adventist Church''). In this case, it is caused by the fact that ``Seventh-day'' is tokenized into three words in TB2.  For the $ORG/MISC \rightarrow O$ examples, ``Guess Who'' is a rock band and ``Sounds Live Feels Live'' is a concert tour by Australian pop-rock band 5 Seconds of Summer. These named entities tend to contain common English verbs with their first letters capitalized. It is difficult to annotate them correctly if the model does not have access to world and domain knowledge. Our analysis points to the future Twitter NER research to introduce text perturbations into training and to encode commonsense knowledge into NER modeling.


\begin{table}[h]
\centering
\begin{tabular}{l|cc|c}
\textbf{Training data} & \textbf{TB2} & \textbf{WNUT17} & \textbf{F1 Drop}\\ \Xhline{2\arrayrulewidth}
spaCy & 52.20 & 44.93 & 7.27$\downarrow$\\
FLAIR &  62.12 & 55.11 &  7.01$\downarrow$\\
\hdashline
HgFace-BERTweet & 73.71 & 59.43 & 14.28$\downarrow$\\
spaCy-BERTweet &  73.79 & 60.77 & 13.02$\downarrow$ \\
\hline
Stanza &  60.14 & 56.40  &  3.74$\downarrow$\\
\end{tabular}
\caption{Comparison among NER models trained on TB2 vs. WNUT17 on TB2 test in entity-level F1. ``HgFace'' stands for ``HuggingFace''.}
\label{tab:NER_WNUT17}
\end{table}

\subsubsection{NER Models Trained on WNUT17}

We train spaCy, FLAIR, Stanza, HuggingFace-BERTwee, and spaCy-BERTweet NER models on the four-class version of WNUT17 and evaluate their performance on the TB2 test. In Table \ref{tab:NER_WNUT17}, we compare the performance of these models trained on WNUT17 against the ones trained on TB2. We show that the performance of all the models drops significantly if we use the pre-trained model from WNUT17, meaning the \texttt{Tweebank-NER} dataset is still challenging for current NER models and can be used as an additional benchmark to evaluate NER models.

\subsection{Performance in Syntactic NLP Tasks}

Apart from NER, we train and evaluate Stanza models for tokenization, lemmatization, POS tagging, and dependency parsing by leveraging TB2 and UD\_English-EWT. For each task, we compare our models against previous work on the TB2 test set. 

\subsubsection{Tokenization Performance}

In Table \ref{tab:tokenization}, we observe that the Stanza model trained on TB2 outperforms Twpipe tokenizer, the previous SOTA model, and it achieves slightly higher performance than the spaCy tokenizer. We also find that blending TB2 and UD\_English-EWT for training brings down the tokenization performance slightly. This is probably because the data source of UD\_English-EWT, which is collected from weblogs, newsgroups, emails, reviews, and Yahoo! Answers, represents a different dialect from Twitter English.

\begin{table}[h]
\centering
\begin{tabular}{l|c}
\textbf{System} & \textbf{F1}\\ \Xhline{2\arrayrulewidth}
Twokenizer& 94.6\\
Stanford CoreNLP& 97.3\\
UDPipe v1.2 & 97.4\\
Twpipe& 98.3\\\hline
spaCy (TB2) & 98.57 \\
spaCy (TB2+EWT) & 95.57 \\
\hline
Stanza (TB2) & \textbf{98.64}\\
Stanza (TB2+EWT)& 98.59 \\

\end{tabular}
\caption{Tokenizer comparison on the TB2 test set. ``TB2'' indicates to use TB2 for training. ``TB2+EWT'' indicates to combine TB2 and UD English-EWT for training. Note that the first four results are rounded to one decimal place by Liu et al., (2018). }
\label{tab:tokenization}
\end{table}

\subsubsection{Lemmatization Performance}

 None of the previous Twitter NLP work reports the lemmatization performance on TB2. As shown in Table \ref{tab:lemmatization}, the Stanza model outperforms the other two rule-based (NLTK and spaCy) and one neural (FLAIR) baseline approaches on TB2. This is not surprising because the Stanza ensemble lemmatizer makes good use of both ruled-based dictionary lookup and seq2seq learning. Similar to what we observe in the tokenization experiments, the combined data setting brings down the performance of FLAIR and Stanza models. 


\begin{table}[h]
\centering
\begin{tabular}{l|c}
\textbf{System} & \textbf{F1} \\ \Xhline{2\arrayrulewidth}
NLTK& 88.23 \\
spaCy & 85.28 \\
Flair (TB2) & 96.18 \\
Flair (TB2+EWT) & 84.54 \\
\hline
Stanza (TB2) & \textbf{\textbf{98.25}}\\
Stanza (TB2+EWT)& 85.45
\end{tabular}
\caption{Lemmatization results on the TB2 test set. ``TB2'' is to use TB2 for training. ``TB2+EWT'' is to combine TB2 and UD English-EWT for training.}
\label{tab:lemmatization}
\end{table}

\subsubsection{POS Tagging Performance}
As shown in Table \ref{tab:POS}, HuggingFace-BERTweet (TB2) replicates the SOTA performance from BERTweet \cite{nguyen2020bertweet} in terms of accuracy. When trained on the combined data of TB2 and UD\_English-EWT, HuggingFace-BERTweet achieves the best performance (95.38\%) in accuracy out of all the models. Compared to HuggingFace-BERTweet, spaCy-transformers models perform worse. The spaCy-XLM-RoBERTa trained on TB2 is 1.3\% lower than \newcite{nguyen2020bertweet}. We conjecture that the difference is mainly due to the implementations of the POS tagging layer between spaCy and HuggingFace-BERTweet, which is the same as \newcite{nguyen2020bertweet}. Among the non-transformer models, Stanza achieves competitive performance compared with \newcite{owoputi2013improved}'s tagger with CRF (93.53\% vs. 94.6\%). Stanza outperforms all other non-transformer baselines including Stanford CoreNLP, spaCy, FLAIR, and \newcite{ma2016end}. Interestingly, we observe that adding UD\_English-EWT for training improves the performance of non-transformer models and HuggingFace-BERTweet but slightly brings down the performance of spaCy-transformers models. 





\begin{table}[!ht]
\centering
\begin{tabular}{l|c}
\textbf{System} & \textbf{UPOS} \\ \Xhline{2\arrayrulewidth}
Stanford CoreNLP& 90.6 \\
\newcite{owoputi2013improved} (greedy) & 93.7 \\
\newcite{owoputi2013improved} (CRF)& 94.6\\
\newcite{ma2016end} & 92.5 \\ 
\hdashline
BERTweet \cite{nguyen2020bertweet} & 95.2 \\
\hline
spaCy (TB2) & 86.72 \\ 
spaCy (TB2+EWT) & 88.84 \\ 
FLAIR (TB2) & 87.85 \\ 
FLAIR (TB2+EWT) & 88.19 \\
\hdashline
HuggingFace-BERTweet (TB2) & 95.21 \\
HuggingFace-BERTweet (TB2+EWT) & \textbf{95.38} \\
spaCy-BERTweet (TB2)  & 87.61 \\
spaCy-BERTweet (TB2+EWT) & 86.31   \\ 
spaCy-XLM-RoBERTa (TB2)  & 93.90  \\
spaCy-XLM-RoBERTa (TB2+EWT) & 93.75  \\  \hline
Stanza (TB2) & 93.20\\
Stanza (TB2+EWT)& 93.53 \\

\end{tabular}
\caption{POS Tagging comparison in accuracy on the TB2
test set. ``TB2'' is to use TB2 for training. ``TB2+EWT'' is to combine TB2 and UD English-EWT for training. Please note that the first five results are rounded to one decimal place by Liu et al., (2018).}
\label{tab:POS}
\end{table}

\subsubsection{Dependency Parsing Performance}

For dependency parsing experiments, spaCy-XLM-RoBERTa (TB2) achieves the SOTA performance (Table \ref{tab:depparse}), surpassing \newcite{liu2018parsing} (Ensemble) by 0.42\% in UAS\footnote{It is difficult to compare their LAS with ours due to the difference in decimal places.}. Besides that, the Stanza parser achieves the same UAS score and has a close LAS score ($-$0.3\%) compared to this best non-transformer performance (UAS 82.1\% + LAS 77.9\%) reported by the distilled parser. As \newcite{liu2018parsing} mentioned, the ensemble model is 20 times larger in size compared to the Stanza parser, although the former performs better. Finally, we confirm that the combination of TB2 and UD\_English-EWT training sets boost the performance for non-transformer models \cite{liu2018parsing}. The data combination brings down the performance of transformer-based models, which is consistent with our observations in tokenization, POS tagging, and dependency parsing.

\begin{table}[h]
\centering
\begin{tabular}{l|cc}
\textbf{System}& \textbf{UAS} & \textbf{LAS} \\ \Xhline{2\arrayrulewidth}
\newcite{kong2014dependency} & 81.4 & 76.9 \\
\newcite{dozat2017stanford} &81.8 & 77.7\\
\newcite{ballesteros2015improved} &80.2 & 75.7 \\ 
\newcite{liu2018parsing} (Ensemble) &83.4 &\textbf{79.4}\\ 
\newcite{liu2018parsing} (Distillation) &82.1 &77.9\\ 
\hline
spaCy (TB2) & 66.93 & 58.79 \\
spaCy (TB2 + EWT) & 72.06 & 63.84 \\
\hdashline
spaCy-BERTweet (TB2)  & 76.32 & 71.72 \\
spaCy-BERTweet (TB2+EWT) & 76.18   & 69.28 \\ 
spaCy-XLM-RoBERTa (TB2)  & \textbf{83.82} & \textbf{79.39} \\
spaCy-XLM-RoBERTa (TB2+EWT) & 81.02   & 75.43 \\ 
\hline
Stanza (TB2) &  79.28 & 74.34 \\
Stanza (TB2 + EWT) &82.10 & 77.60

\end{tabular}
\caption{Dependency parsing comparison on the TB2
test set. ``TB2'' indicates to use TB2 for training. ``TB2+EWT'' indicates to combine TB2 and UD English-EWT for training. Note that the first six results are rounded to one decimal place by Liu et al., (2018).}
\label{tab:depparse}
\end{table}

\section{Conclusion}

In this paper, we introduce four-class named entities to Tweebank V2, a popular Twitter dataset within the Universal Dependencies framework, creating a new NER benchmark called \texttt{Tweebank-NER}. We evaluate our annotations and observe good inter-annotator agreement score in pairwise F1 for NER annotation. We train Twitter-specific NLP models (NER, tokenization, lemmatization, POS tagging, dependency parsing) on the dataset with Stanza and compare our models against existing work and NLP frameworks. Our Stanza models show SOTA performance on tokenization and lemmatization and competitive performance in NER, POS tagging, and dependency parsing on TB2. We also train BERT-based methods to establish a strong benchmark on \texttt{Tweebank-NER} and achieve SOTA performance in POS tagging and dependency parsing on TB2. Finally, we publish our dataset and release the Stanza pipeline \verb|Twitter-Stanza|, which is easy to download and use with Stanza's Python interface. We also release the BERTweet-based NER and POS tagger on Hugging Face Hub. We hope that our research not only contributes annotations to an important dataset but also enables other researchers to use off-the-shelf NLP models for social media analysis. 

\section{Acknowledgements}

We would like to thank Alan Ritter, Yuhui Zhang, Zifan Lin, and anonymous reviewers, who gave precious advice and comments on our paper. We also want to thank John Bauer and Yijia Liu for answering questions related to Stanza and Twpipe. Finally, we would like to thank MIT Center for Constructive Communication for funding our research.

\section{Bibliographical References}\label{reference}

\bibliographystyle{lrec2022-bib}
\bibliography{lrec2022-example}

\section{Language Resource References}
\label{lr:ref}
\bibliographystylelanguageresource{lrec2022-bib}
\bibliographylanguageresource{languageresource}

\end{document}